\documentclass[11pt,a4paper]{article}
\usepackage[utf8]{inputenc}
\usepackage[T1]{fontenc}
\usepackage{amsmath,amssymb}
\usepackage{graphicx}
\usepackage{booktabs}
\usepackage{hyperref}
\usepackage[margin=2.5cm]{geometry}
\usepackage[numbers]{natbib}
\usepackage{xcolor}
\usepackage{multirow}
\usepackage{array}
\usepackage{float}
\usepackage{siunitx} 
\usepackage{caption}
\title{Entity-Aware Sequence Transduction for Player-Centric Ball Action Spotting}

\author{
Ruifeng Wang$^{1}$,
Di Yang$^{2*}$,
Jiangtao Wang$^{2*}$\\[0.5em]
\small $^{1}$School of Artificial Intelligence \& Data Science, USTC, Hefei, China\\
\small $^{2}$Suzhou Institute for Advanced Research, USTC, Suzhou, China\\
\small $^{*}$Correspondence:
di.yang@ustc.edu.cn; wangjiangtao@ustc.edu.cn
}

\date{}

\begin{document}
\maketitle

\begin{abstract}
Player-centric ball action spotting requires temporally precise event detection together with actor attribution in crowded, partially observed multi-agent sports videos. Existing Denoising Sequence Transduction (DST) baselines treat the player-role dimension as part of a flattened frame-level representation, which weakens the inductive bias for modeling player-specific temporal evolution and inter-player interactions. 
To address this limitation, we propose Multi-Entity Denoising Sequence Transduction (ME-DST). ME-DST keeps the role-slot dimension throughout encoding. It uses temporal attention to model the history of each role slot, and spatial attention to exchange information across role slots at each frame. This factorized design gives the model a direct structure for separating within-player evolution from inter-player context. We also add learnable role embeddings, tracking-derived tactical features, and fused visual predictions from X3D-L and Swin3D-S.
Experiments on the FOOTPASS dataset show that ME-DST reaches a Micro F1 of 0.778. This improves the strongest official TAAD+DST baseline by 10.3 percentage points. Controlled ablations show that preserving the entity axis and encoding role identity are central to this gain. These results suggest that explicit entity modeling is an effective inductive bias for player-centric sports event understanding.

\end{abstract}

\section*{Introduction}
Soccer is the most widely followed sport in the world, and modern broadcasts generate large-scale video archives that are valuable for event indexing, tactical analysis, coaching, scouting, and fan engagement. This scale has made automated soccer video understanding an important direction in AI for sport. The task is also technically demanding. Soccer actions are brief, frequent, and often visually subtle, and many events occur in crowded scenes where several players move around the ball at the same time. A useful model must therefore go beyond coarse clip-level recognition. It must localize events in time, distinguish fine-grained ball actions, and identify the player responsible for each action. These requirements make player-centric soccer analysis a natural testbed for structured multi-agent video understanding.

A central task in this setting is action spotting~\cite{giancola2018soccernet}. Unlike conventional action recognition, which operates on pre-trimmed clips, action spotting requires simultaneously predicting the temporal location and class of every event in untrimmed video---discriminating true events from a background that occupies the vast majority of a 90-minute match while maintaining frame-level temporal precision. Ball action spotting~\cite{cioppa2023soccernet} narrows this formulation to events involving contact with the ball. Unlike general action spotting, which targets a broad spectrum of match incidents spanning goals, cards, and set pieces, ball action spotting isolates the dense, rapid exchanges that constitute open play---passes, drives, crosses, shots, headers, throw-ins, tackles, and blocks---offering a finer granularity of analysis. This shift added a significant layer of complexity: models must detect not only salient moments such as shots, but also subtle and fleeting ball contacts that unfold in a fraction of a second yet are critical to the rhythm of the game. Player-centric ball action spotting (SoccerNet 2026 Challenge, \url{https://www.soccer-net.org/challenges/2026}) further raises the bar by requiring the model to identify \textit{who} performed each detected action. Beyond predicting the action class and its temporal location, a model must attribute every event to a specific player, identified by jersey number, and evaluated under joint four-dimensional matching---frame, team, jersey, and class all correct for a true positive. This progression from event-level to actor-level understanding demands qualitatively new capabilities in identity reasoning, role understanding, and fine-grained spatio-temporal modeling.

The SoccerNet benchmark series established action spotting as a
reproducible task with standardized evaluation protocols
~\cite{giancola2018soccernet,deliege2021soccernet}. Early work centered on feature-pooling approaches: NetVLAD++~\cite{giancola2021temporally} aggregated frame-level ResNet features over temporal windows using NetVLAD pooling~\cite{arandjelovic2016netvlad}, achieving 53.4 average-mAP on SoccerNet-v2~\cite{deliege2021soccernet} but suffering from coarse temporal resolution. CALF~\cite{cioppa2020context} introduced direct timestamp regression with a context-aware loss that penalized temporal errors in proportion to their distance from ground-truth, improving tight-mAP by 12.5\% over NetVLAD++. E2E-Spot~\cite{hong2022spotting} replaced the two-stage feature-extraction-and-classification pipeline with an end-to-end RegNetY transformer, reaching 61.8 tight-mAP. Subsequent Transformer-based methods explored complementary directions for soccer action spotting. ASTRA~\cite{xarles2023astra} combined audio-visual modeling with balanced mixup and uncertainty-aware temporal displacement, whereas COMEDIAN~\cite{denize2024comedian} used self-supervised pretraining and knowledge distillation to improve spatial--temporal Transformer initialization. Dense Detection Anchors~\cite{soares2022temporally} reformulated the problem as dense prediction over one anchor per time instant per class, using a Transformer encoder trunk with temporal displacement regression, and won the SoccerNet 2022 Challenge~\cite{giancola2022soccernet} with 67.8 tight-mAP.

Ball action spotting narrowed the focus to ball-contact events. The 2023 SoccerNet Ball Action Spotting Challenge~\cite{cioppa2023soccernet} introduced a 2-class formulation (Pass and Drive); the winning solution combined EfficientNetV2-B0~\cite{tan2021efficientnetv2} with 3D convolutional blocks in a slow-fusion architecture, achieving 87.9 mAP@1, while the runner-up, BME~\cite{wang2023boosted}, applied boosted model ensembling to the E2E-Spot baseline (85.5 mAP@1). The 2024 edition~\cite{cioppa2024soccernet} expanded the taxonomy to 12 classes. T-DEED~\cite{xarles2024tdeed} introduced a temporally discriminative encoder-decoder that sharpened per-token attention distributions for precise localization, winning the challenge. The 2025 Team Ball Action Spotting task further required each detected action to be assigned to the team performing it~\cite{cioppa2025soccernet}. The winning solution extended T-DEED with a unified action--team classification head and large-scale pretraining, achieving 60.03 Team-mAP@1. These methods progressively enrich event-level prediction from action and time to action, time, and team, but they do not identify the individual player responsible for each event.

The transition to player-centric ball action spotting, formalized by the SoccerNet 2026 Challenge, introduces an additional attribution requirement: a correct prediction must match the ground-truth frame, team, jersey number, and action class. FOOTPASS~\cite{ochin2025footpass} subsequently established the player-centric benchmark, providing 54 broadcast matches with synchronized tracking data and 102,992 annotations across eight ball-action classes.

FOOTPASS introduces three baselines that progressively incorporate player and sequence context. The Track-Aware Action Detector (TAAD)~\cite{singh2023large} performs frame-wise visual classification for individual role slots using an X3D-L backbone. TAAD+GNN~\cite{ochin2025gnn} augments these predictions with message passing over a player-proximity graph. The strongest baseline, TAAD+DST~\cite{ochin2025beyondpixels}, formulates event spotting as Denoising Sequence Transduction. Given visual logits and tracking features over a 750-frame window, DST uses a Transformer encoder--decoder to generate an ordered sequence of player-specific events. This sequence-level formulation substantially outperforms frame-wise TAAD, demonstrating the importance of longer-term context and structured event decoding.

The original DST encoder nevertheless removes the explicit entity structure of its input before temporal modeling. At each frame, features from all 26 role slots are concatenated into a single vector, and self-attention is then applied over the resulting temporal sequence. Consequently, the encoder must infer several distinct forms of structure within the same flattened representation: the temporal evolution of each player, the relationships among different players, and the identity associated with each fixed role slot. Although this formulation captures global match context, \textbf{it does not explicitly distinguish within-player temporal dependencies from cross-player interactions}.

\begin{figure*}[htbp]
\centering
\includegraphics[width=0.95\textwidth]{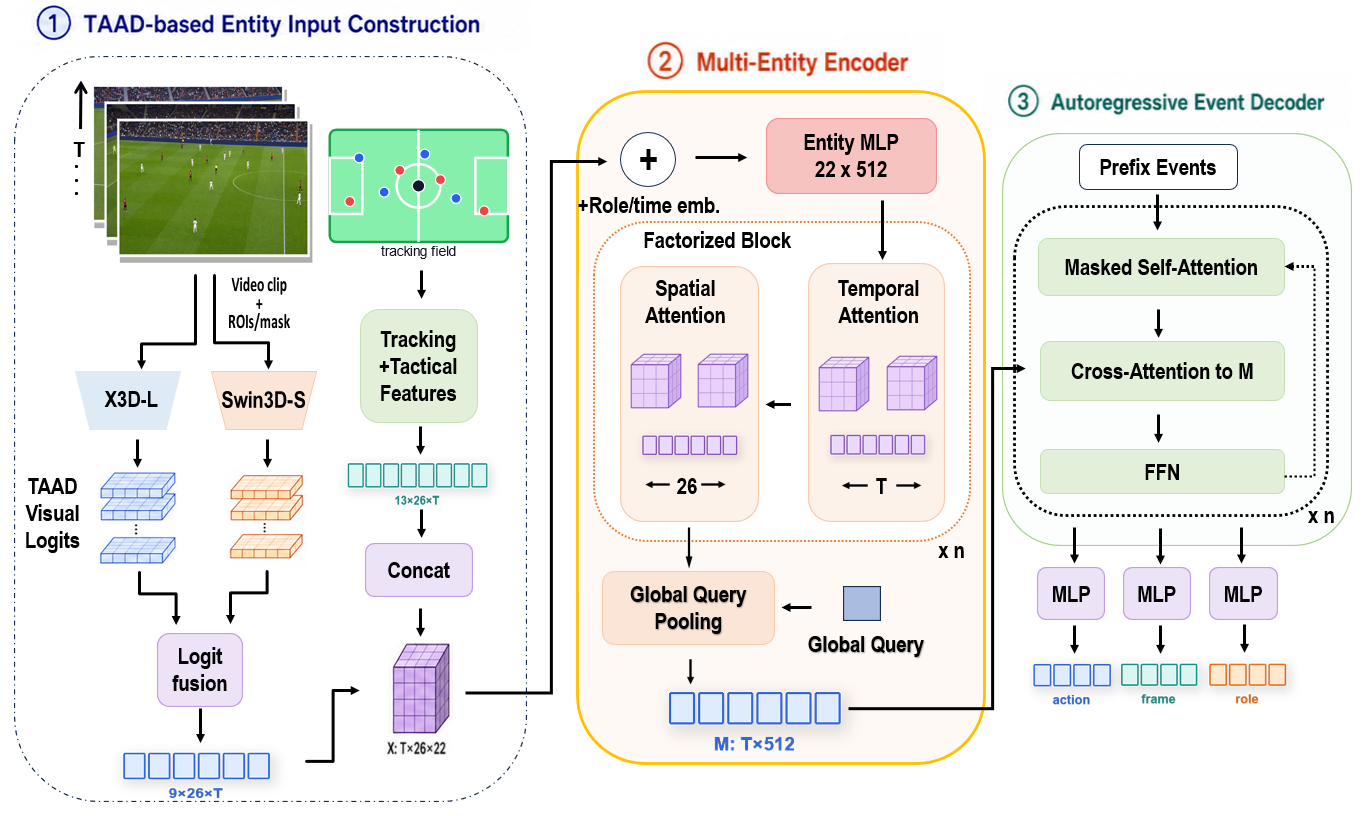}
\small
\caption{Overview of ME-DST. 
(1) Fused TAAD logits from X3D-L and Swin3D-S are concatenated with tracking and tactical features to form the entity tensor $\mathbf{X}\in\mathbb{R}^{T\times26\times22}$. 
(2) Role- and time-aware entity embeddings are processed by factorized temporal and spatial attention and pooled into the memory $\mathbf{M}\in\mathbb{R}^{T\times512}$. 
(3) An autoregressive decoder predicts the action, frame, and role of each player-centric event.}
\label{fig:architecture}
\end{figure*}

We address this limitation with Multi-Entity Denoising Sequence Transduction (ME-DST), an extension of the FOOTPASS DST framework that preserves the role-slot dimension throughout encoding. 
ME-DST, illustrated in Fig.~\ref{fig:architecture}, first models the temporal evolution of each role slot independently and then exchanges information among role slots at each frame. This factorization separates player-specific temporal reasoning from cross-player spatial reasoning while retaining the autoregressive denoising decoder of DST. Learnable role embeddings provide explicit slot identity, allowing the encoder to distinguish representations associated with different tactical roles.

We additionally examine two sources of information that complement the multi-entity encoder. First, tracking data are augmented with tactical descriptors derived from player motion, local proximity, pressure, and position relative to the attacking goal. These features make geometric relations available directly to the sequence model instead of requiring them to be reconstructed exclusively from absolute coordinates. Second, visual predictions from X3D-L~\cite{feichtenhofer2020x3d} and Swin3D-S~\cite{liu2022video} TAAD branches are fused to combine complementary convolutional and transformer-based video representations.

The main contributions of this work are as follows:
\begin{itemize}
    \item We introduce ME-DST, an entity-aware sequence transduction framework for player-centric ball action spotting. ME-DST preserves player-role slots during encoding and predicts an ordered sequence of player-specific events.

    \item We design a factorized multi-entity encoder that separates two forms of reasoning that are mixed in flat DST representations. Temporal attention models the evolution of each role slot over time, while spatial attention models interactions among role slots at each frame. Learnable role embeddings further provide stable role identity, which helps bind temporal evidence to the correct player.

    \item We integrate tracking-derived tactical descriptors and complementary visual predictions into the entity-aware representation. 
    
    \item Extensive experiments on the FOOTPASS dataset show that ME-DST reaches a Micro F1 of 0.778 and improves the strongest official TAAD+DST baseline by 10.3 percentage points.
\end{itemize}

\section*{Results}
This section reports the validation performance of ME-DST. We first compare against the official FOOTPASS baselines, then isolate the effects of the main design choices through controlled ablations. We conclude with per-class analysis and representative examples that illustrate the player-centric nature of the task.
\subsection*{Experimental Setup}

Experiments are conducted on the FOOTPASS dataset~\cite{ochin2025footpass}, which contains 54 full-match broadcasts from major European soccer leagues during the 2023--24 season, including 48 matches for training, 3 for validation, and 3 for testing. The dataset provides 102,992 frame-level annotations covering eight event categories: Drive, Pass, Cross, Throw-in, Shot, Header, Tackle, and Block. Following the official evaluation protocol, performance is measured using micro-averaged precision, recall, and F1. Predictions are regarded as correct if they fall within $\pm12$ frames ($\pm0.48$ s at 25 fps) of the ground-truth event, using a confidence threshold of 0.15.

Visual predictions are generated by two TAAD models with pretrained X3D-L~\cite{feichtenhofer2020x3d} and Swin3D-S~\cite{liu2022video} backbones. Both models process 50-frame clips together with player regions of interest and visibility masks to produce frame-wise logits for 26 role slots. The Swin3D-S predictions are temporally smoothed using a one-dimensional Gaussian filter ($\sigma=1.5$) before fusion. Unless otherwise stated, the visual prior used by subsequent DST models is obtained by weighted averaging of the two prediction streams:
\[
\mathbf{z}=0.6\mathbf{z}_{\mathrm{X3D}}+0.4\mathbf{z}_{\mathrm{Swin}}.
\]

Unless explicitly specified in the corresponding ablation, all experiments use the same optimization strategy and training protocol. Detailed implementation settings are provided in the Methods section.

\subsection*{Main Results}

Table~\ref{tab:main} summarizes the comparison between ME-DST and the official FOOTPASS baselines on the validation set. Unless otherwise noted, the reported ME-DST configuration uses fused visual predictions, eight-dimensional tactical features, two factorized attention blocks, role embeddings, and no global temporal refinement.

\begin{table}[htbp]
\centering
\caption{Comparison with existing methods on the FOOTPASS validation set.}
\label{tab:main}
\begin{tabular}{@{}lccc@{}}
\toprule
\textbf{Method} & \textbf{Micro F1} & \textbf{Precision} & \textbf{Recall} \\
\midrule
TAAD~\cite{ochin2025footpass} & 0.359 & 0.256 & 0.599 \\
TAAD+GNN~\cite{ochin2025footpass} & 0.521 & 0.445 & 0.627 \\
TAAD+DST~\cite{ochin2025footpass} & 0.675 & 0.682 & 0.668 \\
\midrule
\textbf{ME-DST (ours)} & \textbf{0.778} & \textbf{0.792} & \textbf{0.765} \\
\bottomrule
\end{tabular}
\end{table}

ME-DST achieves the best overall performance, reaching a Micro F1 of 0.778 on the FOOTPASS validation set. Compared with the strongest official baseline, TAAD+DST, the proposed method improves Micro F1 by 10.3 percentage points, while increasing precision from 0.682 to 0.792 and recall from 0.668 to 0.765. Larger gains are observed relative to TAAD and TAAD+GNN, with absolute improvements of 41.9 and 25.7 percentage points in Micro F1, respectively.

The baseline results in Table~\ref{tab:main} are taken from the official FOOTPASS benchmark. For the ablation studies presented in the following sections, we use our reproduced X3D-L+DST implementation as the reference model to ensure a consistent experimental setting.

\subsection*{Ablation Study}
\label{sec:ablation}

To evaluate the contribution of individual design choices, we perform ablation experiments on the FOOTPASS validation set following the official SoccerNet 2026 evaluation protocol. Unless otherwise specified, only one component is modified at a time while the remaining settings are kept unchanged.

\subsubsection*{Visual Backbone}

Table~\ref{tab:backbone} reports the effect of replacing the single X3D-L visual branch with fused predictions from X3D-L and Swin3D-S. Without tactical features, visual fusion improves Micro F1 from 0.667 to 0.710. A similar improvement is observed when three-dimensional tactical features are included, increasing Micro F1 from 0.670 to 0.715.

The three-dimensional tactical features provide a modest improvement under both visual backbones. Their effect is limited for the X3D-L model (+0.3 percentage points) and slightly larger for the fused representation (+0.5 percentage points). Among the evaluated configurations in Table~\ref{tab:backbone}, the combination of fused visual predictions and three-dimensional tactical features achieves the highest Micro F1 of 0.715.

\begin{table}[htbp]
\centering
\caption{Comparison of visual backbones with and without tactical features. P = Precision, R = Recall.}
\label{tab:backbone}
\begin{tabular}{@{}lccc@{}}
\toprule
\textbf{Configuration} & \textbf{Micro F1} & \textbf{Precision} & \textbf{Recall} \\
\midrule
X3D-L + DST & 0.667 & 0.678 & 0.656 \\
X3D-L + DST + 3-dim features & 0.670 & 0.684 & 0.656 \\
Fusion + DST & 0.710 & \textbf{0.725} & 0.696 \\
Fusion + DST + 3-dim features & \textbf{0.715} & 0.723 & \textbf{0.708} \\
\bottomrule
\end{tabular}
\end{table}

\subsubsection*{Temporal Preprocessing Module}

We next evaluate whether additional temporal preprocessing improves the sequence representation before the DST encoder. Three alternatives are considered: a one-dimensional convolution (Conv1D), an additional Transformer encoder layer~\cite{vaswani2017attention}, and a bidirectional GRU~\cite{cho2014learning}.

Table~\ref{tab:temporal} shows that the three temporal modules produce comparable performance under the same visual and feature configuration. Using fused visual predictions and three-dimensional tactical features, Conv1D, Transformer, and GRU achieve Micro F1 scores of 0.710, 0.708, and 0.703, respectively. Conv1D gives the highest score among the three variants, although the performance differences remain within 0.7 percentage points.

Across all evaluated configurations, the best result in this group is obtained by combining fused visual predictions with Conv1D, reaching a Micro F1 of 0.714. Overall, the choice of temporal preprocessing has only a modest influence on performance compared with the visual backbone explored in the previous section.

\begin{table}[htbp]
\centering
\caption{Comparison of temporal preprocessing modules.}
\label{tab:temporal}
\begin{tabular}{@{}lccc@{}}
\toprule
\textbf{Configuration} & \textbf{Micro F1} & \textbf{Precision} & \textbf{Recall} \\
\midrule
X3D-L + DST + Conv1D & 0.666 & 0.682 & 0.650 \\
X3D-L + DST + 3-dim features + Conv1D & 0.674 & 0.689 & 0.660 \\
Fusion + DST + Conv1D & \textbf{0.714} & \textbf{0.734} & 0.696 \\
Fusion + DST + 3-dim features + Conv1D & 0.710 & 0.721 & \textbf{0.700} \\
Fusion + DST + 3-dim features + Transformer & 0.708 & 0.718 & 0.698 \\
Fusion + DST + 3-dim features + GRU & 0.703 & 0.725 & 0.683 \\
\bottomrule
\end{tabular}
\end{table}

\subsubsection*{Feature Engineering}

We further examine how the dimensionality of tactical features affects performance under different encoder architectures. Table~\ref{tab:features} compares three feature configurations for the flat DST encoder and two configurations for the multi-entity encoder.

For the flat DST encoder, introducing three-dimensional tactical features increases the Micro F1 from 0.710 to 0.715, whereas replacing them with the full eight-dimensional feature set reduces performance slightly to 0.708. In contrast, the multi-entity encoder benefits from the richer representation: Micro F1 improves from 0.758 with three-dimensional features to 0.778 when the complete eight-dimensional feature set is used.

This pattern suggests that tactical descriptors are not universally beneficial as additional input dimensions, rather, their utility depends on whether the encoder can preserve the player-level structure to which these descriptors refer.

\begin{table}[htbp]
\centering
\caption{Impact of tactical feature dimensionality under flat DST and multi-entity encoders.}
\label{tab:features}
\begin{tabular}{@{}lccc@{}}
\toprule
\textbf{Feature Configuration} & \textbf{Micro F1} & \textbf{Precision} & \textbf{Recall} \\
\midrule
\multicolumn{4}{@{}l}{\textit{Flat DST encoder (Fusion backbone)}} \\
\quad No tactical features & 0.710 & 0.725 & 0.696 \\
\quad 3-dim (speed, teammate dist, opponent dist) & 0.715 & 0.723 & 0.708 \\
\quad 8-dim (full set) & 0.708 & 0.720 & 0.696 \\
\midrule
\multicolumn{4}{@{}l}{\textit{Multi-entity encoder, no global refinement (Fusion backbone)}} \\
\quad 3-dim & 0.758 & 0.769 & 0.748 \\
\quad 8-dim & \textbf{0.778} & \textbf{0.792} & \textbf{0.765} \\
\bottomrule
\end{tabular}
\end{table}

\subsubsection*{Multi-Entity Encoder Components}

Finally, we examine the contribution of individual components in the proposed multi-entity encoder. All configurations in Table~\ref{tab:ablation} use fused visual predictions and the eight-dimensional tactical feature set, while only the encoder architecture is modified.

Among all evaluated components, role embeddings have the largest impact on performance. Removing role embeddings decreases the Micro F1 from 0.736 to 0.569, corresponding to a reduction of 16.7 percentage points. By comparison, removing spatial attention results in a much smaller decrease, from 0.736 to 0.733.

The global temporal refinement layer does not improve performance in our setting. Removing this layer increases the Micro F1 from 0.736 to 0.778, which is the best result obtained in this ablation study. Reducing the encoder depth from two factorized blocks to one yields intermediate performance, with Micro F1 scores of 0.749 and 0.753 for the refined and non-refined variants, respectively.

Taken together, these results indicate that explicit role information is the most influential component of the proposed encoder, whereas the global refinement layer is unnecessary for the final model configuration adopted in this work. 

\begin{table}[h]
\centering
\caption{Ablation of ME-DST encoder components using fused visual logits and eight-dimensional tactical features. Base: 2 blocks, spatial attention, role embeddings, and global refinement (F1 = 0.736).}
\label{tab:ablation}
\begin{tabular}{@{}lccc@{}}
\toprule
\textbf{Configuration} & \textbf{Micro F1} & \textbf{Precision} & \textbf{Recall} \\
\midrule
Base (2 blocks + spatial + role + refine) & 0.736 & 0.757 & 0.717 \\
\quad -- global refinement & \textbf{0.778} & \textbf{0.792} & \textbf{0.765} \\
\quad -- spatial attention & 0.733 & 0.744 & 0.723 \\
\quad -- role embeddings & 0.569 & 0.574 & 0.565 \\
\midrule
1 block + spatial + role & 0.749 & 0.756 & 0.742 \\
1 block + spatial + role, no refine & 0.753 & 0.764 & 0.743 \\
\bottomrule
\end{tabular}
\end{table}

\subsection*{Per-Class Analysis}

Table~\ref{tab:perclass} compares the per-class performance of the reproduced X3D-L+DST baseline and the final ME-DST model. Improvements are observed across all action categories, although their magnitude varies considerably.

\begin{table}[h]
\centering
\caption{Per-class performance: reproduced X3D-L + DST vs.\ ME-DST best configuration (2 blocks + role, no refine). P = Precision, R = Recall, TP = True Positives, FN = False Negatives. $\Delta$F1 denotes the absolute F1 improvement of ME-DST over the reproduced baseline.}
\label{tab:perclass}
\begin{tabular}{@{}lccccccccccc@{}}
\toprule
& \multicolumn{5}{c}{\textbf{X3D-L + DST}} & \multicolumn{5}{c}{\textbf{ME-DST (ours)}} & \\
\cmidrule(lr){2-6} \cmidrule(lr){7-11}
\textbf{Class} & \textbf{P} & \textbf{R} & \textbf{F1} & \textbf{TP} & \textbf{FN} & \textbf{P} & \textbf{R} & \textbf{F1} & \textbf{TP} & \textbf{FN} & \textbf{$\Delta$F1} \\
\midrule
Drive  & 0.659 & 0.651 & 0.655 & 1609 & 861  & 0.793 & 0.776 & 0.784 & 1916 & 554  & +0.129 \\
Pass   & 0.725 & 0.703 & 0.714 & 2152 & 907  & 0.822 & 0.795 & 0.808 & 2431 & 628  & +0.094 \\
Cross  & 0.539 & 0.495 & 0.516 & 55   & 56   & 0.716 & 0.748 & 0.732 & 83   & 28   & +0.216 \\
Throw-in & 0.735 & 0.742 & 0.738 & 72   & 25   & 0.755 & 0.732 & 0.743 & 71   & 26   & +0.005 \\
Shot   & 0.623 & 0.493 & 0.550 & 33   & 34   & 0.667 & 0.597 & 0.630 & 40   & 27   & +0.080 \\
Header & 0.303 & 0.272 & 0.287 & 44   & 118  & 0.548 & 0.457 & 0.498 & 74   & 88   & +0.211 \\
Tackle & 0.000 & 0.000 & 0.000 & 0    & 26   & 0.053 & 0.038 & 0.044 & 1    & 25   & +0.044 \\
Block  & 0.262 & 0.205 & 0.230 & 16   & 62   & 0.423 & 0.385 & 0.403 & 30   & 48   & +0.173 \\
\midrule
Overall & 0.678 & 0.656 & 0.667 & 3981 & 2089 & 0.792 & 0.765 & 0.778 & 4646 & 1424 & +0.111 \\
\bottomrule
\end{tabular}
\end{table}

\begin{figure*}[htbp]
\centering
\includegraphics[width=0.95\textwidth]{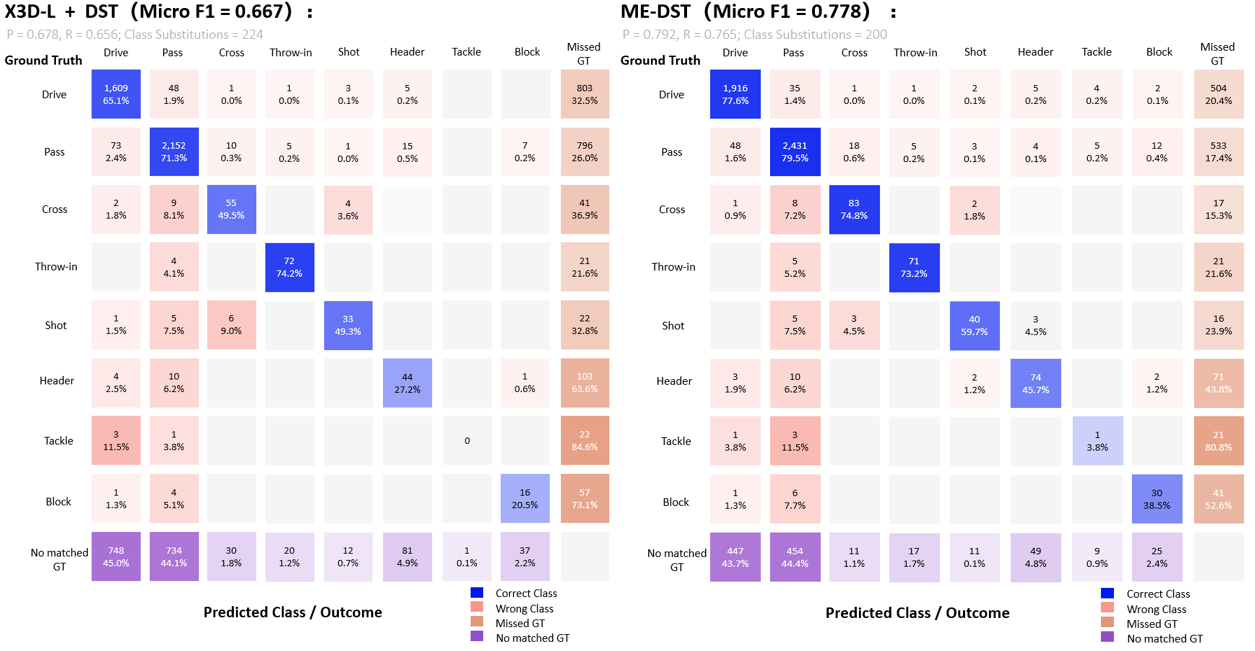}
\caption{Event-level confusion matrices for the reproduced X3D-L+DST baseline and ME-DST. Rows denote ground-truth action classes and columns denote predicted action classes. Diagonal cells are official class-aware true positives. Off-diagonal cells show residual player-aware temporal matches with an incorrect predicted class. The rightmost column reports missed ground-truth events, and the bottom row reports predictions with no matched ground-truth event. Percentages are row-normalized.}
\label{fig:perclass_detection_matrix}
\end{figure*}

Fig.~\ref{fig:perclass_detection_matrix} complements Table~\ref{tab:perclass} with an event-level confusion matrix. To remain consistent with the official spotting protocol, diagonal entries are first obtained using class-aware matching over frame, team, jersey number, and action class. The remaining unmatched predictions and ground-truth events are then paired within the same player and temporal window while ignoring the action class, exposing residual class substitutions as off-diagonal entries. Under this analysis, ME-DST reduces class substitutions from 224 to 200, while also decreasing residual missed events and unmatched predictions.

The largest gains are observed for Cross (+0.216), Header (+0.211), and Block (+0.173), while Drive and Pass also improve substantially by 0.129 and 0.094, respectively. In contrast, Throw-in changes only marginally (+0.005). Although Tackle remains the most challenging category, the proposed model detects one event that is missed by the reproduced baseline, increasing the class F1 from 0 to 0.044.

The same trend is observed under different evaluation subsets. Compared with the Fusion+DST baseline, ME-DST improves recall from 0.760 to 0.826 on ball-visible events and from 0.391 to 0.480 on ball-hidden events. Recall also increases on replay segments (0.474 to 0.513) and live-play segments (0.720 to 0.794). The improvement is particularly pronounced under ball-hidden conditions, where recall increases by 8.9 percentage points.

\begin{figure*}[htbp]
\centering
\includegraphics[width=0.95\textwidth]{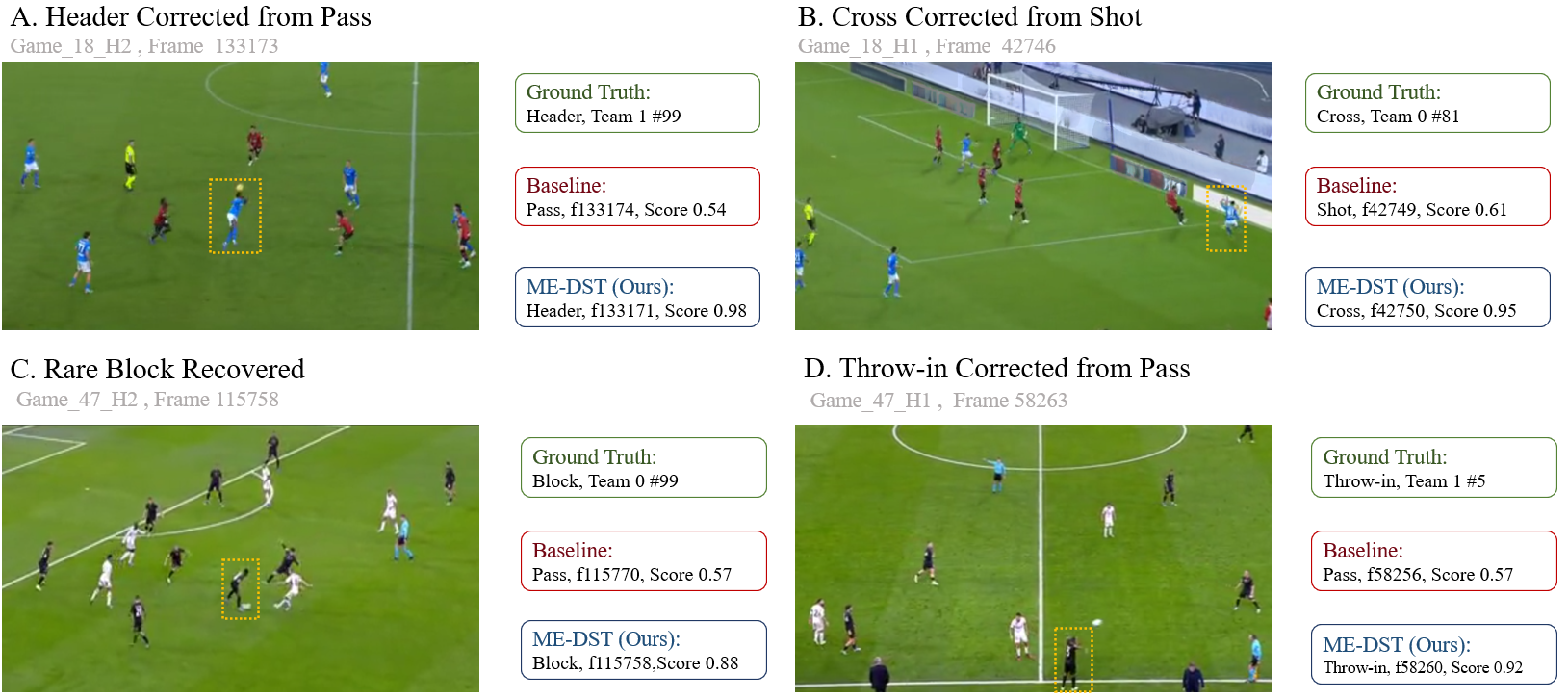}
\small
\caption{Baseline errors corrected by ME-DST. Yellow dashed boxes mark the annotated actor; green, red, and blue boxes show the ground truth, reproduced X3D-L+DST baseline, and ME-DST results, respectively. (A) A Header is corrected from pass by ME-DST. (B) A Cross mistaken as a Shot is corrected (C) A rare Block is discovered by ME-DST. (D) A Throw-in by the line is corrected from pass.}
\label{fig:examples}
\end{figure*}

Fig.~\ref{fig:examples} shows representative validation examples. These examples illustrate the joint time-action-player matching required by player-centric ball action spotting. ME-DST produces predictions that more closely match the ground truth in these examples, which is consistent with the quantitative improvements reported above.

\section*{Discussion}
The results suggest that player-centric ball action spotting benefits from keeping the player-role axis explicit during sequence modeling. This section discusses the implications of this finding, the role of explicit identity cues, and the main limitations of the current approach.
\subsection*{Factorized Attention for Multi-Agent Sports Understanding}

Our experiments indicate that explicitly separating temporal modeling from inter-player interaction provides a more effective representation for multi-agent sports understanding than the flat sequence formulation adopted by the original DST encoder. This conclusion is supported by the consistent improvements observed across the main benchmark, the component ablations, and the per-class analysis. Similar factorized designs have recently been explored for American football defensive coverage analysis~\cite{song2026decoding} and badminton stroke recognition~\cite{ibh2023tempose}, suggesting that separating temporal dynamics from entity interaction may constitute a broadly applicable design principle for structured sports analysis.

A possible explanation is that the two types of dependencies captured in soccer videos differ in nature. The temporal evolution of an individual player provides evidence for predicting player-specific actions, whereas many events are also influenced by the spatial configuration of surrounding teammates and opponents~\cite{gavrilyuk2020actor,li2021groupformer,
yuan2021agentformer,mohamed2020social}. Modeling these dependencies separately allows temporal information to be preserved within each player trajectory before information is exchanged across players through spatial attention. By contrast, the original DST encoder flattens all player representations into a single sequence, requiring temporal dynamics and inter-player relationships to be learned simultaneously within the same representation.

The feature engineering experiments provide further evidence that model architecture and handcrafted tactical cues should be considered jointly rather than independently. Increasing the feature dimensionality produces little benefit for the flat DST encoder, whereas the same feature set consistently improves performance in the multi-entity formulation. This observation suggests that richer tactical descriptors become more useful when the underlying representation explicitly preserves player-level structure throughout the encoding process.

Several limitations should also be acknowledged. Although the ablation study isolates the contribution of individual components within the proposed framework, it does not fully disentangle architectural changes from differences in model capacity. Additional controlled comparisons, such as parameter-matched variants or alternative multi-agent encoders, would provide a more rigorous assessment of the proposed design. We leave this direction for future work.

\subsection*{The Importance of Explicit Role Information}

Among all architectural components evaluated in this study, removing role embeddings results in the largest performance degradation, reducing the Micro F1 from 0.736 to 0.569. This finding suggests that explicitly encoding player roles provides information that cannot be recovered reliably from motion trajectories alone.

One possible explanation is that players occupying different tactical roles exhibit distinct spatial distributions and action patterns over the course of a match. Providing this information explicitly enables the encoder to distinguish players with similar short-term trajectories but different tactical responsibilities. In contrast, when role embeddings are removed, the model must infer these distinctions solely from observed motion, which is likely to be more challenging, particularly under limited training data.

Although our implementation represents each role using a learnable embedding vector, the underlying idea is not specific to soccer. Many multi-agent sports involve relatively stable player identities or tactical assignments, suggesting that explicit role information may provide a useful inductive bias beyond the FOOTPASS benchmark. Evaluating this hypothesis on additional sports datasets represents an interesting direction for future work.

\subsection*{When Less Is More: Revisiting Global Temporal Refinement}

Removing the global temporal refinement layer consistently improves performance, increasing the Micro F1 from 0.736 to 0.778. This observation suggests that, under the proposed architecture, the two factorized attention blocks already provide sufficient temporal modeling before the final prediction stage. Each block captures long-range temporal dependencies for individual player representations while simultaneously modeling interactions across players, reducing the need for an additional global Transformer encoder.

One possible interpretation is that the benefit of additional temporal refinement depends on the information already encoded by earlier stages of the network. In our implementation, each player representation has already aggregated information over approximately 750 frames (about 30 seconds of game time) before reaching the refinement layer. Under this setting, an additional round of global temporal attention appears to provide limited benefit and may even interfere with temporal localization. Although this hypothesis is consistent with our ablation results, further analyses, such as attention visualization or boundary localization experiments, would be required to verify the underlying mechanism.

More broadly, these findings suggest that temporal modeling capacity should be considered at the level of the entire architecture rather than individual modules. When long-range temporal dependencies have already been explicitly modeled upstream, adding further temporal refinement may introduce redundancy instead of complementary information. This observation may also be relevant to other sequence modeling problems in which hierarchical temporal representations are employed.

\subsection*{Limitations and Future Directions}

Several limitations qualify the conclusions of this study. First, performance remains strongly affected by the class distribution of FOOTPASS. ME-DST improves all eight action categories, but \textit{Tackle} remains difficult, with only one correctly detected event and an F1 score of 0.044. The F1 scores for \textit{Header} and \textit{Block} also remain below 0.5 despite substantial improvements over the reproduced baseline. These results indicate that modifying the sequence encoder does not by itself resolve the limited supervision available for rare and visually ambiguous actions~\cite{santra2025precise}. Imbalance-aware objectives, targeted sampling, and class-specific temporal augmentation warrant further investigation.

Second, ME-DST operates on visual logits produced by separately trained TAAD branches. Errors in player detection, action classification, or temporal localization are therefore propagated to the sequence model and cannot be corrected through joint visual--sequence optimization. This dependency is particularly relevant under occlusion, where the visual evidence for both the player and the ball may be incomplete. Jointly training the visual branches and the denoising sequence model could help determine how much of the remaining error arises from upstream perception and how much arises from sequence transduction.

The current role representation also depends on the organization and annotations provided by FOOTPASS. Role indices are converted to player identities using shirt-number observations propagated across missing frames. This procedure can be unreliable near substitutions, changes in tactical assignment, or extended periods without a visible shirt number. Similarly, the tactical descriptors are manually specified and rely on normalized tracking coordinates. Learned trajectory representations or identity-association modules may provide greater robustness when role definitions, tracking quality, or camera configurations differ from those in FOOTPASS~\cite{cabado2024beyond}.

Our architectural conclusions should also be interpreted within the scope of the reported ablations. The experiments show that the proposed multi-entity encoder, role embeddings, tactical features, and visual fusion are effective in the evaluated configurations, but they do not fully separate architectural effects from differences in parameter count or optimization. In particular, the benefit of preserving entity structure should be examined using parameter-matched flat and multi-entity encoders, multiple random seeds, and additional player-centric datasets. The limited number of validation matches further makes uncertainty estimates important for future comparisons.

Beyond action spotting, Denoising Sequence Transduction could be extended to settings in which the output remains an ordered event sequence but the prediction target changes, such as action anticipation~\cite{dalal2025action} or tactical event forecasting. These directions would test whether explicit entity-wise encoding remains useful when the model must predict future rather than observed events. Evaluation on other sports would further clarify which aspects of ME-DST depend on soccer-specific role structure and which transfer to more general multi-agent sequence transduction.

In summary, our results identify a specific limitation of the original FOOTPASS DST encoder: concatenating all role slots before temporal modeling removes the explicit player dimension from the encoded sequence. Preserving this dimension and separately modeling within-player temporal dynamics and cross-player interactions improves player-centric ball action spotting in the FOOTPASS setting. ME-DST should therefore be viewed as an entity-aware extension of Denoising Sequence Transduction rather than as a replacement for its sequence-decoding formulation. Establishing how broadly this extension transfers will require controlled evaluation beyond the present benchmark.

\section*{Methods}
This section describes the full ME-DST pipeline. We first define the player-centric ball action spotting task and the event representation used by the decoder. We then describe the visual logits, tracking-derived tactical features, multi-entity encoder, and autoregressive decoding procedure. Finally, we provide the training, inference, and evaluation settings used for all experiments.
\subsection*{Problem Formulation}
\label{sec:problem_formulation}
We formulate soccer action spotting as a sequence prediction problem over structured player representations. The input consists of a soccer broadcast containing $T$ video frames sampled at 25 fps together with synchronized player tracking data organized into 26 predefined team-role slots (13 roles for each team).

The goal is to predict a sequence of ball action events,
\[
\mathcal{E}=\{e_1,e_2,\ldots,e_K\},
\]
where each event is represented as
\[
e_k=(t_k,\ell_k,j_k,c_k,s_k).
\]
Here, $t_k$ denotes the event frame, $\ell_k\in\{0,1\}$ identifies the team, $j_k$ is the jersey number, $c_k$ is one of the eight action categories (Drive, Pass, Cross, Throw-in, Shot, Header, Tackle, or Block), and $s_k$ denotes the prediction confidence.

The decoder predicts an autoregressive sequence over the eight action classes augmented with start-of-sequence (SOS) and end-of-sequence (EOS) tokens. No explicit background token is included in the output vocabulary.

\subsection*{Visual Feature Extraction}
\label{sec:visual_features}

Visual observations are represented using two complementary TAAD branches based on pretrained X3D-L~\cite{feichtenhofer2020x3d} and Swin3D-S~\cite{liu2022video} backbones. Both branches follow the same TAAD detection pipeline, in which hierarchical visual features are aggregated through a feature pyramid, pooled within player regions of interest using ROIAlign, and decoded into frame-wise action logits. The two models differ only in the underlying video backbone, allowing them to capture complementary visual representations while sharing an identical detection framework.

Each branch processes 50-frame clips with a spatial resolution of $352\times640$ and predicts logits in $\mathbb{R}^{9\times26\times T}$, corresponding to nine action categories (eight actions and one background class) for every role slot at each frame. Before fusion, the Swin3D-S predictions are temporally smoothed using a one-dimensional Gaussian filter ($\sigma=1.5$). The final visual representation is obtained by weighted averaging:
\begin{equation}
\mathbf{Z}_{\mathrm{fused}}
=
0.6\mathbf{Z}_{\mathrm{X3D}}
+
0.4\mathbf{Z}_{\mathrm{Swin}}.
\end{equation}

For each role slot, the fused visual logits are concatenated with 13 tracking-derived features, including five raw tracking attributes and eight engineered tactical descriptors (Section~\ref{sec:features}), producing a 22-dimensional representation that serves as the input to the multi-entity encoder.

\subsection*{Tactical Feature Representation}
\label{sec:features}

In addition to visual predictions, each player is represented by tracking-derived features describing motion, local interactions, and field geometry. The raw tracking data provide five basic attributes at each frame: normalized position $(x,y)$, velocity $(v_x,v_y)$, and a binary visibility indicator.

To enrich the player representation, we compute eight engineered tactical features that can be grouped into three categories.

\textbf{Motion features} describe the player's instantaneous movement and include the speed
\[
f_0=\sqrt{v_x^2+v_y^2},
\]
and the acceleration magnitude
\[
f_3=\sqrt{(\nabla v_x)^2+(\nabla v_y)^2},
\]
computed from consecutive visible observations.

\textbf{Interaction features} characterize the local player configuration. These include the distance to the nearest visible teammate ($f_1$), the distance to the nearest visible opponent ($f_2$), and a pressure index ($f_7$), defined as the normalized number of opponents within a radius of 0.05 in pitch coordinates.

\textbf{Field-geometry features} encode the player's position relative to the attacking goal. They consist of the Euclidean distance to the goal center ($f_4$) together with the sine and cosine of the goal direction angle ($f_5$ and $f_6$), providing a continuous representation of player orientation while avoiding angular discontinuities.

The resulting 13-dimensional tracking representation (five raw and eight engineered features) is concatenated with the fused 9-dimensional visual logits to produce a 22-dimensional representation for each role slot.

For the feature ablation experiments, the three-dimensional configuration retains only speed, nearest-teammate distance, and nearest-opponent distance, whereas the full configuration includes all eight engineered tactical features.

\subsection*{Multi-Entity DST Architecture}
\label{sec:me_dst}

The proposed Multi-Entity DST (ME-DST) encoder transforms the per-player representations described above into a global event representation through four successive stages: player embedding, factorized temporal--spatial encoding, global aggregation, and autoregressive decoding.

\paragraph*{Player representation.}

For each role slot and each frame, the 22-dimensional input vector is independently projected into a $d_{\mathrm{model}}=512$ latent representation using a two-layer multilayer perceptron with GELU activation and LayerNorm,
\begin{equation}
\mathbf{h}^{(0)}_{t,n}
=
\mathrm{MLP}(\mathbf{x}_{t,n}),
\qquad
\mathbf{h}^{(0)}_{t,n}\in\mathbb{R}^{512},
\end{equation}
where $t$ denotes the frame index and $n$ denotes the role slot. A learnable role embedding is added to every player representation,
\begin{equation}
\mathbf{h}^{(0)}_{t,n}
\leftarrow
\mathbf{h}^{(0)}_{t,n}
+
\mathbf{r}_n,
\end{equation}
followed by standard sinusoidal positional encoding computed from frame indices within the current temporal window~\cite{vaswani2017attention}.

\paragraph*{Factorized temporal--spatial encoding.}

The encoded player representations are processed by two identical factorized attention blocks. Within each block, temporal self-attention is first applied independently to each player trajectory,
\begin{equation}
\mathbf{H}^{\mathrm{temp}}_n
=
\mathrm{TemporalAttn}
(\mathbf{H}^{(i-1)}_n),
\end{equation}
after which spatial self-attention models interactions among all role slots at every frame,
\begin{equation}
\mathbf{H}^{\mathrm{spat}}_t
=
\mathrm{SpatialAttn}
(\mathbf{H}^{\mathrm{temp}}_t).
\end{equation}
Both attention modules are implemented using Pre-LN Transformer encoder layers with eight attention heads and a hidden dimension of 512. Spatial attention is evaluated in chunks to reduce memory consumption without changing the computation.

\paragraph*{Global event representation.}

The player-level representations are aggregated into a shared event representation using a learnable query token,
\begin{equation}
\mathbf{M}_t
=
\mathrm{LayerNorm}
\big(
\mathrm{MHA}
(\mathbf{q},
\mathbf{H}^{\mathrm{spat}}_t,
\mathbf{H}^{\mathrm{spat}}_t)
+
\mathbf{q}
\big).
\end{equation}

An additional Transformer encoder operating along the temporal dimension can optionally refine this global memory. As shown in the ablation study, the best-performing configuration omits this refinement stage.

\paragraph*{Autoregressive event decoding.}

The final event representation is decoded using a six-layer Pre-LN Transformer decoder~\cite{xiong2020layer}. At each decoding step, the model predicts the action category, player role, and event frame jointly through three output heads. Decoding terminates when the EOS token is generated or when the maximum decoding length of 25 events is reached.

\subsection*{Training}

The model is trained by jointly optimizing action classification, player-role prediction, and temporal localization. The overall objective is the sum of three cross-entropy losses,
\begin{equation}
\mathcal{L}
=
\mathcal{L}_{\mathrm{action}}
+
\mathcal{L}_{\mathrm{role}}
+
\mathcal{L}_{\mathrm{frame}}.
\end{equation}

Label smoothing ($\epsilon=0.05$)~\cite{szegedy2016rethinking} is applied to the action and role prediction losses, whereas the frame localization loss is optimized without smoothing. To alleviate class imbalance, the action classification loss uses class-specific weights, while all padding positions are excluded from loss computation.

Optimization is performed using AdamW~\cite{kingma2014adam} with linear learning-rate warmup followed by exponential decay. Gradient checkpointing~\cite{chen2016training} is enabled for the multi-entity encoder to reduce memory consumption during training. Unless otherwise specified, all experiments reported in this work use the same optimization settings.

\subsection*{Inference}
\label{sec:inference}
During inference, each match half is partitioned into consecutive 750-frame windows, with zero-padding applied only to the final incomplete window. For each window, the autoregressive decoder predicts an ordered sequence of candidate events and terminates when either an EOS token is generated or the maximum decoding length is reached.

Each predicted event consists of an action category, a role index, a frame location, and an associated confidence score. The predicted role index is subsequently converted to the corresponding team side and jersey number using the available shirt-number annotations with temporal interpolation across missing observations. Events assigned to the neutral role or localized outside the valid temporal range are discarded.

The remaining predictions are merged across all windows to produce the final event sequence for evaluation.

\subsection*{Evaluation Metrics}

Evaluation follows the official SoccerNet 2026 Challenge protocol (\url{https://www.soccer-net.org/challenges/2026}). Predictions with confidence scores below 0.15 are discarded before evaluation. A predicted event is considered correct if it can be uniquely matched to a ground-truth event of the same action class, team side, and jersey number within a temporal tolerance of $\pm12$ frames. Each ground-truth event can be matched to at most one prediction.

Performance is reported using micro-averaged precision, recall, and F1 over all evaluated matches.

\section*{Data Availability}

The FOOTPASS dataset~\cite{ochin2025footpass} is available through the SoccerNet organization on Hugging Face (\url{https://huggingface.co/SoccerNet}) under the access conditions specified by its maintainers. Access to the broadcast videos requires acceptance of the SoccerNet Non-Disclosure Agreement. The fused X3D-L and Swin3D-S logits and the evaluation outputs generated in this study are available from the corresponding author upon reasonable request and are not part of the official FOOTPASS release.

\section*{Code Availability}

The code used to train and evaluate the models in this study is available at the ME-DST GitHub repository (\url{https://github.com/WRF32-10/ME-DST}).

\section*{Author Contributions}

R. W. was responsible for conceiving the study, developing the method, implementing the models, conducting the experiments, analysing the results, and drafting the manuscript. D. Y. and J. W. supervised the study and revised the manuscript. All authors read and approved the final manuscript.

\section*{Acknowledgements}
This work was supported by the National Natural Science Foundation of China under Grant No. 62502492 and the Natural Science Foundation of Jiangsu Province Basic Research Program under Grant No. BK20250489.

\section*{Competing Interests}

All authors declare no financial or non-financial competing interests.

\subsection*{Use of generative AI}

Large language models were used to assist with language editing, LaTeX formatting, and bibliographic consistency checks. All scientific content, experimental design, analyses, references, and conclusions were verified and approved by the authors.



\begin{thebibliography}{99}

\bibitem{giancola2018soccernet}
Giancola, S., Amine, M., Dghaily, T. \& Ghanem, B. SoccerNet: A scalable dataset for action spotting in soccer videos. In \textit{Proc. IEEE/CVF Conf. Comput. Vis. Pattern Recognit. Workshops} 1711--1721 (2018).

\bibitem{cioppa2023soccernet}
Cioppa, A. \textit{et al.} SoccerNet 2023 challenges results. \textit{Sports Eng.} \textbf{27}, 24; 10.1007/s12283-024-00466-4 (2024).

\bibitem{deliege2021soccernet}
Deli\`ege, A. \textit{et al.} SoccerNet-v2: A dataset and benchmarks for holistic understanding of broadcast soccer videos. In \textit{Proc. IEEE/CVF Conf. Comput. Vis. Pattern Recognit. Workshops} (2021).

\bibitem{giancola2021temporally}
Giancola, S. \& Ghanem, B. Temporally-aware feature pooling for action spotting in soccer broadcasts. In \textit{Proc. IEEE/CVF Conf. Comput. Vis. Pattern Recognit. Workshops} 4490--4499 (2021).

\bibitem{arandjelovic2016netvlad}
Arandjelovi\'c, R., Gronat, P., Torii, A., Pajdla, T. \& Sivic, J. NetVLAD: CNN architecture for weakly supervised place recognition. In \textit{Proc. IEEE Conf. Comput. Vis. Pattern Recognit.} 5297--5307 (2016).

\bibitem{cioppa2020context}
Cioppa, A. \textit{et al.} A context-aware loss function for action spotting in soccer videos. In \textit{Proc. IEEE/CVF Conf. Comput. Vis. Pattern Recognit.} (2020).

\bibitem{hong2022spotting}
Hong, J., Zhang, H., Gharbi, M., Fisher, M. \& Fatahalian, K. Spotting temporally precise, fine-grained events in video. In \textit{Proc. Eur. Conf. Comput. Vis.} 33--51 (2022).

\bibitem{xarles2023astra}
Xarles, A., Escalera, S., Moeslund, T. B. \& Clap\'es, A. ASTRA: An action spotting TRAnsformer for soccer videos. In \textit{Proc. 6th Int. Workshop Multimedia Content Anal. Sports} 93--102 (2023).

\bibitem{denize2024comedian}
Denize, J., Liashuha, M., Rabarisoa, J., Orcesi, A. \& H\'erault, R. COMEDIAN: Self-supervised learning and knowledge distillation for action spotting using transformers. In \textit{Proc. IEEE/CVF Winter Conf. Appl. Comput. Vis. Workshops} 530--540 (2024).

\bibitem{soares2022temporally}
Soares, J. V. B., Shah, A. \& Biswas, T. Temporally precise action spotting in soccer videos using dense detection anchors. In \textit{Proc. IEEE Int. Conf. Image Process.} 2796--2800; 10.1109/ICIP46576.2022.9897256 (2022).

\bibitem{giancola2022soccernet}
Giancola, S. \textit{et al.} SoccerNet 2022 challenges results. In \textit{Proc. 5th Int. ACM Workshop Multimedia Content Anal. Sports} 75--86; 10.1145/3552437.3558545 (2022).

\bibitem{tan2021efficientnetv2}
Tan, M. \& Le, Q. V. EfficientNetV2: Smaller models and faster training. In \textit{Proc. 38th Int. Conf. Mach. Learn.} 10096--10106 (2021).

\bibitem{wang2023boosted}
Wang, L., Guo, H. \& Liu, B. A boosted model ensembling approach to ball action spotting in videos: The runner-up solution to the CVPR 2023 SoccerNet Challenge. \textit{arXiv} 2306.05772 (2023).

\bibitem{cioppa2024soccernet}
Cioppa, A. \textit{et al.} SoccerNet 2024 challenges results. \textit{arXiv} 2409.10587 (2024).

\bibitem{xarles2024tdeed}
Xarles, A., Escalera, S., Moeslund, T. B. \& Clap\'es, A. T-DEED: Temporal-discriminability enhancer encoder-decoder for precise event spotting in sports videos. In \textit{Proc. IEEE/CVF Conf. Comput. Vis. Pattern Recognit. Workshops} (2024).

\bibitem{cioppa2025soccernet}
Giancola, S. \textit{et al.} SoccerNet 2025 challenges results. \textit{arXiv} 2508.19182 (2025).

\bibitem{ochin2025footpass}
Ochin, J., Chekroun, R., Stanciulescu, B. \& Manitsaris, S. FOOTPASS: A multi-modal multi-agent tactical context dataset for play-by-play action spotting in soccer broadcast videos. \textit{arXiv} 2511.16183 (2025).

\bibitem{singh2023large}
Singh, G., Choutas, V., Saha, S., Yu, F. \& Van Gool, L. Spatio-temporal action detection under large motion. In \textit{Proc. IEEE/CVF Winter Conf. Appl. Comput. Vis.} 6009--6018 (2023).

\bibitem{ochin2025gnn}
Ochin, J., Devineau, G., Stanciulescu, B. \& Manitsaris, S. Game state and spatio-temporal action detection in soccer using graph neural networks and 3D convolutional networks. In \textit{Proc. 14th Int. Conf. Pattern Recognit. Appl. Methods} 636--646 (2025).

\bibitem{ochin2025beyondpixels}
Ochin, J., Chekroun, R., Stanciulescu, B. \& Manitsaris, S. Beyond pixels: Leveraging the language of soccer to improve spatio-temporal action detection in broadcast videos. In \textit{Advanced Concepts for Intelligent Vision Systems: 22nd International Conference, ACIVS 2025} 552--563; 10.1007/978-3-032-07343-3\_44 (Springer, 2026).

\bibitem{feichtenhofer2020x3d}
Feichtenhofer, C. X3D: Expanding architectures for efficient video recognition. In \textit{Proc. IEEE/CVF Conf. Comput. Vis. Pattern Recognit.} 200--210 (2020).

\bibitem{liu2022video}
Liu, Z. \textit{et al.} Video Swin Transformer. In \textit{Proc. IEEE/CVF Conf. Comput. Vis. Pattern Recognit.} 3202--3211 (2022).

\bibitem{vaswani2017attention}
Vaswani, A. \textit{et al.} Attention is all you need. \textit{Adv. Neural Inf. Process. Syst.} \textbf{30} (2017).

\bibitem{cho2014learning}
Cho, K. \textit{et al.} Learning phrase representations using RNN encoder--decoder for statistical machine translation. In \textit{Proc. 2014 Conf. Empir. Methods Nat. Lang. Process.} 1724--1734 (2014).

\bibitem{song2026decoding}
Song, K. \textit{et al.} Decoding defensive coverage responsibilities in American football using factorized attention-based transformer models. In \textit{Sports Analytics: Third International Conference, ISACE 2026} 170--187 (Springer, 2026).

\bibitem{ibh2023tempose}
Ibh, M., Grasshof, S., Witzner, D. \& Madeleine, P. TemPose: A new skeleton-based transformer model designed for fine-grained motion recognition in badminton. In \textit{Proc. IEEE/CVF Conf. Comput. Vis. Pattern Recognit. Workshops} (2023).

\bibitem{gavrilyuk2020actor}
Gavrilyuk, K., Sanford, R., Javan, M. \& Snoek, C. G. M. Actor-transformers for group activity recognition. In \textit{Proc. IEEE/CVF Conf. Comput. Vis. Pattern Recognit.} 839--848 (2020).

\bibitem{li2021groupformer}
Li, S. \textit{et al.} GroupFormer: Group activity recognition with clustered spatial-temporal transformer. In \textit{Proc. IEEE/CVF Int. Conf. Comput. Vis.} 13668--13677 (2021).

\bibitem{yuan2021agentformer}
Yuan, Y., Weng, X., Ou, Y. \& Kitani, K. M. AgentFormer: Agent-aware transformers for socio-temporal multi-agent forecasting. In \textit{Proc. IEEE/CVF Int. Conf. Comput. Vis.} 9813--9823 (2021).

\bibitem{mohamed2020social}
Mohamed, A., Qian, K., Elhoseiny, M. \& Claudel, C. Social-STGCNN: A social spatio-temporal graph convolutional neural network for human trajectory prediction. In \textit{Proc. IEEE/CVF Conf. Comput. Vis. Pattern Recognit.} 14424--14432 (2020).

\bibitem{santra2025precise}
Santra, S., Chudasama, V., Wasnik, P. \& Balasubramanian, V. N. Precise event spotting in sports videos: Solving long-range dependency and class imbalance. In \textit{Proc. IEEE/CVF Conf. Comput. Vis. Pattern Recognit.} 3163--3172 (2025).

\bibitem{cabado2024beyond}
Cabado, B. \textit{et al.} Beyond the Premier: Assessing action spotting transfer capability across diverse domains. In \textit{Proc. IEEE/CVF Conf. Comput. Vis. Pattern Recognit. Workshops} 3386--3398 (2024).

\bibitem{dalal2025action}
Dalal, A. \textit{et al.} Action anticipation from SoccerNet football video broadcasts. In \textit{Proc. IEEE/CVF Conf. Comput. Vis. Pattern Recognit. Workshops} (2025).

\bibitem{xiong2020layer}
Xiong, R. \textit{et al.} On layer normalization in the Transformer architecture. In \textit{Proc. Int. Conf. Mach. Learn.} (2020).

\bibitem{szegedy2016rethinking}
Szegedy, C., Vanhoucke, V., Ioffe, S., Shlens, J. \& Wojna, Z. Rethinking the Inception architecture for computer vision. In \textit{Proc. IEEE Conf. Comput. Vis. Pattern Recognit.} (2016).

\bibitem{kingma2014adam}
Kingma, D. P. \& Ba, J. Adam: A method for stochastic optimization. In \textit{Proc. 3rd Int. Conf. Learn. Represent.} (2015).

\bibitem{chen2016training}
Chen, T., Xu, B., Zhang, C. \& Guestrin, C. Training deep nets with sublinear memory cost. \textit{arXiv} 1604.06174 (2016).

\end{thebibliography}
\end{document}